\documentclass[conference]{IEEEtran}
\IEEEoverridecommandlockouts
\usepackage{times}
\usepackage{soul}
\usepackage{url}
\usepackage[hidelinks]{hyperref}
\usepackage[utf8]{inputenc}
\usepackage[small]{caption}
\usepackage{graphicx}
\usepackage{amsmath}
\usepackage{amsthm}
\usepackage{booktabs}
\usepackage{algorithm}
\usepackage{algorithmic}
\usepackage[switch]{lineno}
\usepackage[export]{adjustbox}

\usepackage{cite}
\usepackage{amsmath,amssymb,amsfonts}
\usepackage{algorithmic}
\usepackage{graphicx}
\usepackage{textcomp}
\usepackage{xcolor}
\def\BibTeX{{\rm B\kern-.05em{\sc i\kern-.025em b}\kern-.08em
    T\kern-.1667em\lower.7ex\hbox{E}\kern-.125emX}}
\begin{document}

\title{  Improving the Reusability of Pre-trained Language Models in Real-world Applications}
\author{\IEEEauthorblockN{Somayeh Ghanbarzadeh\textsuperscript{*}, Hamid Palangi\textsuperscript{$\dagger$}, Yan Huang\textsuperscript{*}, Radames Cruz Moreno\textsuperscript{$\dagger$}, and Hamed Khanpour\textsuperscript{$\dagger$} }
\IEEEauthorblockA{\textit{\textsuperscript{*}Department of Computer Science and Engineering } \\
\textit{University of North Texas, Denton, TX}\\
Emails: somayehghanbarzadeh@my.unt.edu, yan.huang@unt.edu}
\IEEEauthorblockA{\textit{$\dagger$ Microsoft Research, Redmond, WA } \\
Emails:\{hpalangi, radames.cruz, hamed.khanpour\}@microsoft.com}
}

\maketitle

\begin{abstract}
The reusability of state-of-the-art Pre-trained Language Models (PLMs) is often limited by their generalization problem, where their performance drastically decreases when evaluated on examples that differ from the training dataset, known as Out-of-Distribution (OOD)/unseen examples. This limitation arises from PLMs' reliance on spurious correlations, which work well for frequent example types but not for general examples. To address this issue, we propose a training approach called \textbf{Mask-tuning}, which integrates Masked Language Modeling (MLM) training objectives into the fine-tuning process to enhance PLMs' generalization. Comprehensive experiments demonstrate that Mask-tuning surpasses current state-of-the-art techniques and enhances PLMs' generalization on OOD datasets while improving their performance on in-distribution datasets. The findings suggest that Mask-tuning improves the reusability of PLMs on unseen data, making them more practical and effective for real-world applications. 
\end{abstract}


\section{Introduction}
Fine-tuning large-scale Pre-trained Language Models (PLMs) on a specific-task dataset has achieved state-of-the-art performance on a variety of Natural Language Processing (NLP) tasks \cite{b1, b2,b3}.
However, recent studies \cite{b4, b5, b6} have shown that PLMs trained on the large corpora learn spurious correlations, which are prediction patterns that work for frequent example types drawn from the same distribution as the training examples but break down in more challenging cases such as Out-of-Distribution (OOD)/unseen  datasets. If such patterns often yield correct outputs, the fine-tuning training loss function that heavily relies on these patterns cannot incentivize the language model to learn the linguistic patterns from the less frequent example types and generalize them to OOD examples. The inability to generalize patterns to OOD datasets is a major obstacle for PLMs' practical use in real-world applications.  

To improve the PLMs' reusability on unseen data, several solutions have been proposed such as: transfer learning through continuing pre-training (e.g., \cite{b7, b8, b9}) or performing extra fine-tuning (e.g., \cite{b10, b11}) on the domain of interest. However, besides pre-training being computationally expensive, the effectiveness of transfer learning is highly limited to the size of the data, the similarity of the source/target domains, and task complexity \cite{b12}. 
Other studies \cite{b13, b14, b15, b16, b17, b18} proposed different learning strategies focusing on challenging keywords, called data biases, that lead to forming spurious correlations. Although these solutions achieved promising results, their main weakness is the strong assumption of knowing the datasets' biases in advance. Also, since some of these studies \cite{b13, b17} used data augmentation for their proposed learning method,  determining the number of examples required for the best performance is challenging \cite{b19}. Some of these solutions also (e.g.,\cite{b14, b16}) decrease the PLMs' performance on in-distribution datasets.

\begin{figure}[t]
\centering
     \includegraphics[width=0.8\linewidth]{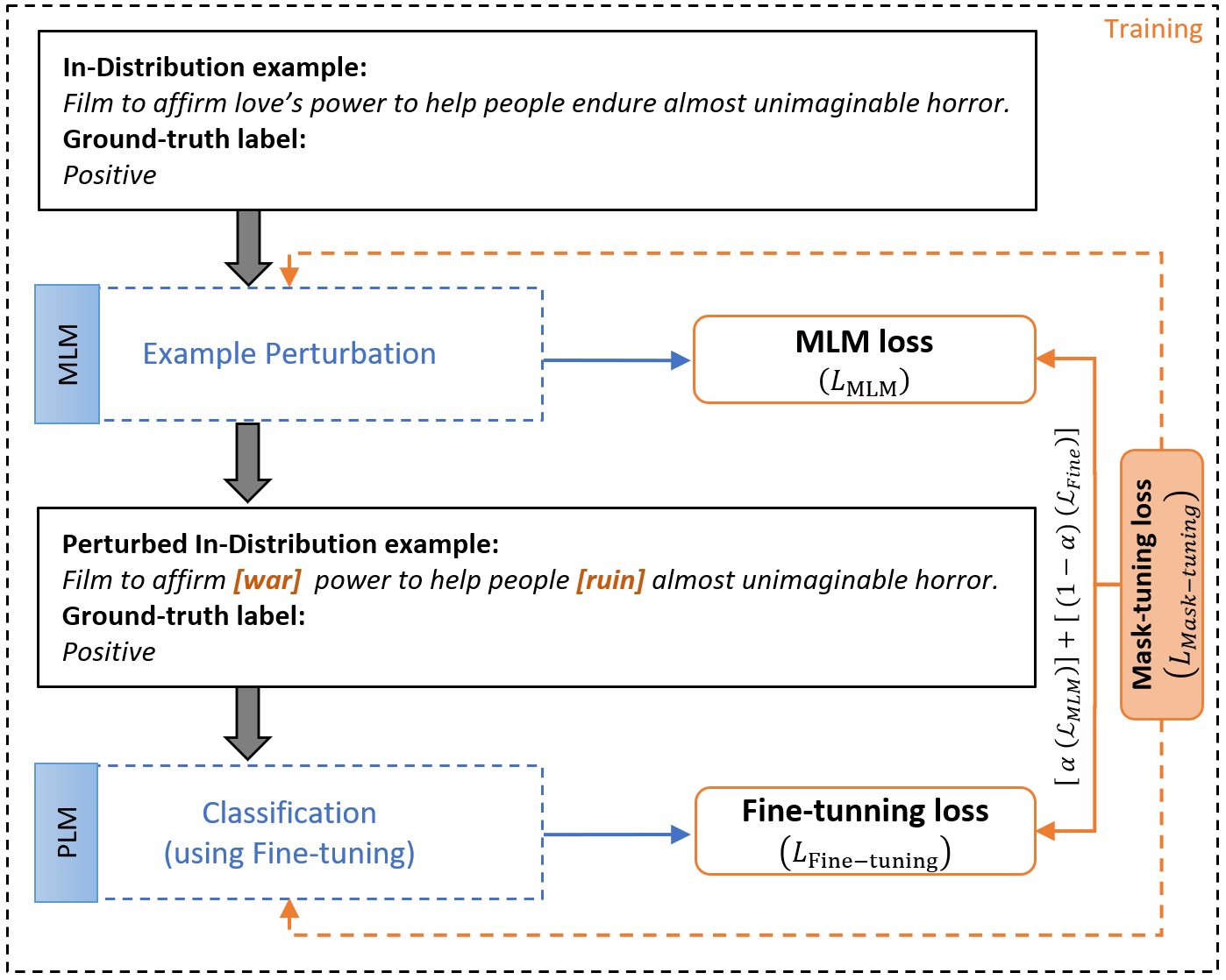}
   
\caption{Illustration of the Mask-tuning's training process. The input of the fine-tuning is only a perturbed version of the in-distribution training example. }
\label{fig:fig1}
\end{figure}

In this study, we propose a novel approach called Mask-tuning to improve the reusability of PLMs on unseen data by enhancing the fine-tuning training process. We draw inspiration from recent data augmentation techniques  (e.g., \cite{b20, b21, b22}) that utilize the Masked Language Model (MLM) to increase the diversity of examples in the training dataset. Unlike those traditional data augmentation approaches, Mask-tuning integrates the MLM training objective into fine-tuning's training process. This unique integrated training method in each training batch primarily perturbs the original training examples using MLM to interrupt the frequent example types (patterns) in the training dataset and generates perturbed examples. Then Mask-tuning classifies the perturbed examples through fine-tuning according to the original examples' ground-truth labels. An integrated loss from perturbation and classification trains the Mask-tuning (Figure \ref{fig:fig1}). Our analysis shows that Mask-tuning creates three times more diversified examples than MLM (Section\ref{sub:D}), demonstrating its effectiveness in enhancing PLMs' generalization.
In summary, our contributions are as follows:\\
\textbf{I.} We study the impact of integrating the MLM training objectives into the fine-tuning for mitigating  PLMs' generalization problem. Our proposed method, Mask-tuning, solely uses the downstream task's training dataset and is a  plug-and-play tool for any PLM that works with original fine-tuning.\\
\textbf{II.} We conduct comprehensive experiments under a consistent evaluation process to verify the effectiveness of Mask-tuning using BERT, RoBERTa, and autoregressive BART language models. We show that Mask-tuning outperforms six state-of-the-art baselines on three downstream tasks' in-distribution and corresponding five OOD datasets. \\
\textbf{III.} We also conduct three ablation studies to compare Mask-tuning performance with data augmentation, Mask-tuning without integrated loss, and sequential training. The results demonstrate the effectiveness of each component of the Mask-tuning for improving the PLMs' generalization.

\section{Related Works} 
\textbf{Transfer Learning;}
Transfer learning for improving the PLMs' generalization has been implemented through two different approaches,  domain- and task-adaptation. The domain-adaptation continues pre-training on examples with the same domain as the downstream task dataset. Some studies proposed to pre-train the language models in their domain of interest from scratch (e.g., \cite{b23, b24}). In contrast, other studies suggested running a second round of pre-training on the domain of interest  (e.g., \cite{b25, b26}). Besides, task-adaptation focuses on sharing knowledge across different tasks \cite{b27, b28}. In task-adaptation, the pre-trained model first fine-tunes on a dataset most related to the target task and then fine-tunes again on the target task. Both approaches have shown the benefits of transfer learning for improving PLM's generalization problem, especially when target training data is limited. However, re-running the pre-training phase on a large-scale dataset is computationally expensive. Furthermore, the effectiveness of transfer learning is highly limited to the size of the data, the similarity between the source and target tasks and domains, and task complexity \cite{b28}. However, same as original fine-tuning, Mask-tuning solely uses the target downstream task dataset.\\ 
\textbf{Data Augmentation;}
Data augmentation is the most straightforward approach for improving the PLMs' robustness. In this method, the downstream task dataset is enriched with examples from the target distribution using various techniques such as: increasing the size of the training dataset \cite{b22, b29}, balancing the existing cues \cite{b30},  adversarial training \cite{b5, b20, b21}, augmenting the standard training dataset with syntactic information  \cite{b31, b32}, creating a partial view to augment the training data \cite{b17}, or dropping span of hidden information \cite{b13}. These approaches generate new examples following different methods and extend the size of the training dataset. Then they use the enlarged dataset for fine-tuning. However, Yoo and Qi\cite{b19} showed that the size of the required data for the best performance is challenging. The main advantages of Mask-tuning over traditional data augmentation is that Mask-tuning benefits from both MLM and fine-tuning objectives. This unique approach allows Mask-tuning to generate perturbed examples that are more diverse and effective for improving the model's generalization. Furthermore, Mask-tuning's perturbed examples are double-validated through fine-tuning classification, ensuring that only high-quality examples are incorporated into the training process. \\
\textbf{Learning Process Improvement; }
Another area of study to enhance the pre-trained models' generalization is introducing new learning techniques that are robust against spurious patterns in the training dataset  \cite{ b33, b14, b34, b30, b16, b35, b18, b37}. These techniques focus on challenging examples or keywords that do not allow the pre-trained models to make shortcuts during training. Thus, they need to design processes to recognize and handle these patterns and keywords in the training examples and re-training the model on extra text corpora. These changes often make the model more complicated and computationally expensive. However, the main weakness of these methods is their strong assumption of knowing these challenging examples or keywords in advance \cite{b12}.

\section{Proposed Method}
We propose Mask-tuning for improving the PLMs' generalization by enhancing the training process of the fine-tuning on downstream tasks' datasets. Mask-tuning perturbs the sequence of words in a training example to interrupt the frequent example types in the training dataset and simultaneously validates the perturbed example according to the original example's label (ground-truth label). For this aim,  Mask-tuning integrates two training objectives: 1) self-supervised Masked Language Modeling (MLM) training objective 
and 2) Fine-tuning's classification function.
In each training batch, Mask-tuning works as follows:

Mask-tuning uses self-supervised MLM to randomly mask a certain percentage of the input tokens in each training example. The MLM training objective is to predict those masked ones based only on their context with a mean cross-entropy loss. We denote the first training phase's loss as MLM-loss ($\mathcal{L}_{MLM}$). The training examples with the predicted token(s), called \emph{perturbed examples}, are fed into fine-tuning to be classified based on the ground-truth label ($y$).  
So, $p_{\theta}(y'=y| \hat{x})$ is the fine-tuning classification function to predict the perturbed example's label ($y'$) based on the perturbed example ($\hat{x}$)  and compute the fine-tuning training loss ($\mathcal{L}_{Fine-tuning}$), where $\theta$ is the PLMs' parameters for the fine-tuning. A weighted aggregation of these two training processes computes Mask-tuning loss ($\mathcal{L}_{Mask-tuning}$) as follows: 
\begin{equation}
\mathcal{L}_{Mask-tuning} = \alpha \  \mathcal{L}_{MLM} + \:(1-\alpha)  \mathcal{L}_{Fine-tuning}
\label{eq:1}
\end{equation}\\ 
where ${\alpha}$ is a weighting factor that is employed to adjust the contribution of each training phase loss for computing the Mask-tuning loss.
The Mask-tuning training objective aims to minimize both self-supervised MLM loss and supervised fine-tuning classification loss. In the following, we present how Mask-tuning benefits from the aggregated loss in each training batch:\\ 
\textbf{I. The MLM's token-prediction output is correct, so MLM training loss is close to zero} ($L_{MLM}$ $\approx$ $0$)\textbf{;}  
In this case, there would be one of the following scenarios: \\ 
1) The fine-tuning classification's output is also correct, so fine-tuning loss is close to zero ($L_{Fine-tuning}$ $\approx$ $0$). In this case, the Mask-tuning loss is also close to zero ($L_{Mask-tuning}$ $\approx$ $0$). For instance, we observed the following example in the third training epoch: \\
$x$: "take care is nicely performed by a quintet of actresses", label ($y$): 1 $\Rightarrow$  $\hat{x}$: "take care is nicely performed by a [\textbf{group}] of actresses", (label ($\hat{y}$): 1 AND prediction: correct).\\
It shows that fine-tuning classification validates that the perturbed example has the same label as the original example. Thus, there is no need for further training. This scenario is the primary goal of the Mask-tuning training objective that differs Mask-tuning from sequential training and data augmentation, which are investigated in our ablation studies (Section \ref{sec:VI}).     \\ 
2) The output of fine-tuning classification is incorrect, so fine-tuning loss is significant ($L_{Fine-tuning}$ $>$ $0$). For instance, we observed the following example from the sentiment analysis task:\\
$x$: "mumbles his way through the movie.", label ($y$): $0$ $\Rightarrow$  $\hat{x}$: "[\textbf{cheer}] his way through the movie .", (label ($\hat{y}$): 1, AND prediction: incorrect).\\
It means fine-tuning classifier did not validate that the perturbed example has the same label as the original example. In this case, the aggregated loss (Mask-tuning loss) is large enough for continuing training to minimize the fine-tuning loss.\\ 
\textbf{II. The MLM's token-prediction output is incorrect, so MLM training loss is significant ($L_{MLM}$ $>$ $0$ )}; Suppose the predicted token(s) is unrelated to the original token(s), and the prediction output is incorrect, and MLM-loss is significant. For instance, we observed the following example in the third training epoch:\\
$x$: "he slaps together his own brand of liberalism", label ($y$): $0$ $\Rightarrow$  $\hat{x}$: "he [\textbf{accept}] together his own brand of liberalism", (label ($\hat{y}$): 1 AND prediction output: incorrect).\\
In this case, the Mask-tuning loss is large enough, regardless of the fine-tuning classification output, to continue training to minimize both  MLM and Fine-tuning losses. 
As we can see,  Mask-tuning is trained based on the perturbed examples  validated by fine-tuning classifier according to the original examples' ground-truth label. For this objective, the aggregated loss incentivizes the MLM and fine-tuning for further training to reach a correct output from both training objectives and minimize their losses (i.e., MLM training loss and fine-tuning loss).



\section{Experimental setup}
To evaluate the effectiveness of our proposed method, we first trained Mask-tuning on the in-distribution downstream tasks' training dataset. Then, we evaluated Mask-tuning on the in-distribution and corresponding OOD evaluation sets. OOD datasets are drawn from a different distribution with more linguistic variation than the in-distribution training set \cite{b6, b5}. PLMs' performance on OOD datasets indicates their reusability on real-world applications.\\
\textbf{Baselines.}
We compared the performance of our proposed approach with the state-of-the-art baselines, which shed light on the existing training challenges in PLMs. These baselines proposed new learning disciplines to improve the PLMs' generalization. For a fair comparison, we followed their evaluation process (e.g., using the same tasks, OOD datasets, and PLMs). \\
\textbf{In-Distribution and OOD Tasks and Datasets.}
In this study, we conducted comprehensive experiments on three tasks from GLUE
 \footnote{https://gluebenchmark.com/tasks} benchmark
 \cite{b37} that have available and wide-used OOD dataset including: Stanford Sentiment Treebank (SST-2), Natural Language
Inference (MNLI \cite{b38}), and Paraphrase Identification (QQP \cite{b39}). Their corresponding OOD datasets are IMDB-Cont \cite{b40} and IMDB-CAD \cite{b4}, HANS \cite{b5}  and AdvNLI \cite{b41}, and PAWS \cite{b6}, respectively.\\ 
\begin{table*}[t]
\caption{ Performance comparison of Mask-tuning (Ours) and various baselines using different PLMs on in-distribution and OOD datasets. All models are trained on  in-distribution dataset and then evaluated on related test set of in-distribution and OOD dataset.The last row also shows the results of a Large Language Model (LLM), GPT-3 Few-shot Prompting \cite{b44} that is discussed in Section \ref{sub:C}.  The Fine-tuning has been implemented using the code from huggingface \cite{b42}.}
\centering
\resizebox{16cm}{!}{

\begin{tabular}{l l lll|cc c|cc}
\hline
\multicolumn{1}{c}{\textbf{}} &\multicolumn{1}{c}{\textbf{}}& \multicolumn{3}{c}{Sentiment} & \multicolumn{3}{c}{NLI} & \multicolumn{2}{c}{Paraphrase}  \\ \hline
\multicolumn{1}{l}{\textbf{}} & \multicolumn{1}{l}{}  &\multicolumn{1}{c|}{In-Dis} &\multicolumn{2}{c|}{OOD}  &\multicolumn{1}{c|}{In-Dis}& \multicolumn{2}{c|}{OOD} &\multicolumn{1}{c|}{In-Dis}& OOD \\ \hline
\multicolumn{1}{l}{\textbf{Model}} & \multicolumn{1}{l|}{\textbf{Baseline}}  &\multicolumn{1}{c|}{\textbf{SST-2}} &\textbf{IMDB-Cont.}    & \textbf{IMDB-CAD}   &\multicolumn{1}{c|}{\textbf{MNLI (m/mm)}}& \textbf{HANS}       &\textbf{ AdvNLI}       &\multicolumn{1}{c|}{\textbf{QQP}}& \textbf{PAWS} \\ \hline
\multicolumn{1}{l}{BERT$_{base}$}  &  \multicolumn{1}{l|}{Fine-tuning}   &\multicolumn{1}{c}{92.43}& \multicolumn{1}{c}{79.08} & \multicolumn{1}{c|}{87.00} & \multicolumn{1}{c}{84.30/83.40}& \multicolumn{1}{c}{56.90}& \multicolumn{1}{c|}{24.12} &\multicolumn{1}{c}{90.80}& \multicolumn{1}{c}{32.80}\\
 &  \multicolumn{1}{l|}{Learned-Mixin+H \cite{b14}}&\multicolumn{1}{c}{-}  & \multicolumn{1}{c}{-}&\multicolumn{1}{c|}{-}& \multicolumn{1}{c}{83.97/-}&66.15  &- &\multicolumn{1}{c}{-}&-\\
     & \multicolumn{1}{l|}{PoE \cite{b16}}&\multicolumn{1}{c}{-}  & \multicolumn{1}{c}{-}&\multicolumn{1}{c|}{-}& \multicolumn{1}{c}{84.19/-}&66.31& -&\multicolumn{1}{c}{\textbf{}}& -\\
 & \multicolumn{1}{l|}{Regularized-conf \cite{b18}}&\multicolumn{1}{c}{-}  & \multicolumn{1}{c}{-}&\multicolumn{1}{c|}{-}&\multicolumn{1}{c}{84.30/84.80}&69.10 & - &\multicolumn{1}{c}{91.50}&39.80\\
  & \multicolumn{1}{l|}{IPT-standard \cite{b32}} &\multicolumn{1}{c}{-}&\multicolumn{1}{c}{-}&\multicolumn{1}{c|}{-}&\multicolumn{1}{c}{84.40/-}& 56.70& - &\multicolumn{1}{c}{-}& -\\\hline
  &\multicolumn{1}{l|}{ Mask-tuning (ours)}  &\multicolumn{1}{c}{\textbf{93.11}$\pm${0.1}}&\textbf{82.75}$\pm${0.3}& \textbf{88.32}$\pm${0.1}& \multicolumn{1}{c}{\textbf{84.75}$\pm${0.2}/\textbf{85.10}$\pm${0.1}}&\textbf{69.52}$\pm${0.2}&\textbf{26.32}$\pm${0.3} &\multicolumn{1}{c}{\textbf{91.54}$\pm${0.1}}&\textbf{46.74}$\pm${0.5}\\
 \hline
 \multicolumn{1}{l}{RoBERTa$_{base}$}     & \multicolumn{1}{l|}{ Fine-tuning }&\multicolumn{1}{c}{94.49}&\multicolumn{1}{c}{84.50} & \multicolumn{1}{c|}{88.40} & \multicolumn{1}{c}{87.60/87.50}&\multicolumn{1}{c}{67.80} & \multicolumn{1}{c|}{31.20}           &\multicolumn{1}{c}{91.50}& \multicolumn{1}{c}{38.45}   \\ 
\multicolumn{1}{l}{}     & \multicolumn{1}{l|}{Span Cutoff \cite{b17}} & \multicolumn{1}{c}{95.40}&\multicolumn{1}{c}{85.50}& \multicolumn{1}{c|}{89.20} &\multicolumn{1}{c}{88.40/-}& \multicolumn{1}{c}{68.40} & \multicolumn{1}{c|}{31.10}           &\multicolumn{1}{c}{92.00}& \multicolumn{1}{c}{38.80}   \\ 
\multicolumn{1}{l}{}     & \multicolumn{1}{l|}{HiddenCut \cite{b13}} &\multicolumn{1}{c}{95.80} &\multicolumn{1}{c}{87.80} & \multicolumn{1}{c|}{90.40} & \multicolumn{1}{c}{88.20/-}&\multicolumn{1}{c}{71.20} & \multicolumn{1}{c|}{32.80}           & \multicolumn{1}{c}{92.00}&\multicolumn{1}{c}{41.50}   \\ 
\multicolumn{1}{l}{}     & \multicolumn{1}{l|}{IPT-standard \cite{b32}}&\multicolumn{1}{c}{-} &\multicolumn{1}{c}{-}  & \multicolumn{1}{c|}{-} & \multicolumn{1}{c}{87.70/-}&\multicolumn{1}{c}{66.30} & \multicolumn{1}{c|}{-}   & \multicolumn{1}{c}{-}&\multicolumn{1}{c}{-}   \\ \hline
\multicolumn{1}{l}{}     & \multicolumn{1}{l|}{{Mask-Tuning (ours) }}&\multicolumn{1}{c}{\textbf{94.60}$\pm${0.1}}    &\multicolumn{1}{c}{\textbf{88.50}$\pm${0.2}}             & \multicolumn{1}{c|}{\textbf{91.62}$\pm${0.1}} &\multicolumn{1}{c}{\textbf{87.72}$\pm${0.1}/\textbf{87.83}$\pm${0.2}}& \multicolumn{1}{c}{\textbf{75.70}$\pm$0.2}       & \multicolumn{1}{c|}{\textbf{37.40}$\pm${0.6}}           &\multicolumn{1}{c}{91.62$\pm${0.2}}& \multicolumn{1}{c}{\textbf{44.37}{$\pm$0.4}}  \\ \hline
 \multicolumn{1}{l}{BART$_{base}$}  &  \multicolumn{1}{l|}{Fine-tuning} &\multicolumn{1}{c}{93.23}  &\multicolumn{1}{c}{82.48} & \multicolumn{1}{c|}{86.03}&\multicolumn{1}{c}{84.60/84.80}&\multicolumn{1}{c}{56.30}& \multicolumn{1}{c|}{30.51} &\multicolumn{1}{c}{90.50}&\multicolumn{1}{c}{32.27} \\
  \multicolumn{1}{l}{}  &  \multicolumn{1}{l|}{Mask-Tuning (ours)} &\multicolumn{1}{c}{\textbf{93.80}$\pm${0.2}}  &\multicolumn{1}{c}{\textbf{84.00}$\pm${0.1}} & \multicolumn{1}{c|}{\textbf{87.83}$\pm${0.3}}&\multicolumn{1}{c}{\textbf{86.08}$\pm${0.2}/\textbf{86.12}$\pm${0.3}}&\multicolumn{1}{c}{\textbf{70.48}$\pm${0.5}}& \multicolumn{1}{c|}{\textbf{35.31}$\pm${0.4}} &\multicolumn{1}{c}{\textbf{91.03}$\pm${0.1}}&\multicolumn{1}{c}{\textbf{45.71}$\pm${0.5}} \\\hline
GPT-3   &  \multicolumn{1}{l|}{ Few-shots Prompting  \cite{b44}}&\multicolumn{1}{c}{-}  & \multicolumn{1}{c}{-}&\multicolumn{1}{c|}{-}& \multicolumn{1}{c}{{77.60*}/-}&75.30  &- &\multicolumn{1}{c}{83.5*}&73.7\\
  \hline
\end{tabular}}
\label{tab:table2}
\end{table*}
\begin{table}[t]
\centering
\caption{Some perturbed examples generated by Mask-tuning.}
\scalebox{1.05}{
\begin{tabular}{cl  }
\hline
1&\multicolumn{1}{l }{"manipulate feminist empowerment tale thinly!"} \\
&\multicolumn{1}{l}{ epoch 1:  [thinly] $\Rightarrow$ [thinly]} \\
&\multicolumn{1}{l }{ epoch 2:  [thinly] $\Rightarrow$ [lyrically]}\\
&\multicolumn{1}{l}{ epoch 3:  [thinly] $\Rightarrow$ [tenderly]}\\
2&\multicolumn{1}{l }{"Convey a strong sense of the girls' environment."} \\
&\multicolumn{1}{l }{ epoch 1:  [strong] $\Rightarrow$ [strong]} \\
&\multicolumn{1}{l }{ epoch 2:  [sense] $\Rightarrow$ [understanding]}\\
&\multicolumn{1}{l }{ epoch 3:  [girls'] $\Rightarrow$ [UNK]}\\
3&\multicolumn{1}{l }{"The lead roles are more than competent."} \\
&\multicolumn{1}{l }{ epoch 1:  [roles] $\Rightarrow$ [roles]} \\
&\multicolumn{1}{l }{ epoch 2:  [competent] $\Rightarrow$ [qualified]}\\
&\multicolumn{1}{l }{ epoch 3:  [roles] $\Rightarrow$ [actors]}\\\hline
\end{tabular}}
\label{tab:table4}
\end{table}
\textbf{Experimental Setup}
Following the baselines, we performed Mask-tuning using two  PLMs: BERT \cite{b1}, and RoBERTa \cite{b3}. Also, we used BART \cite{b2} as an autoregressive language model with a different architecture from BERT.  
Since we have a base and large size of each pre-trained language model, recent study \cite{b43} illustrated that compared with the base size language models, the large size does not necessarily mitigate the PLMs' generalization issue. Besides, the base size models are more efficient and deployable to a broader computational environment and applications. Thus, we chose the base size of these models to show the efficacy of the Mask-tuning method.

We followed the original language models' parameters\footnote{https://github.com/huggingface/transformers} for the two training phases of Mask-tuning. However, we empirically changed the learning rate and batch size for the fine-tuning phase of the Mask-tuning. After several trial runs, the batch size sets to $32$ for double sentence datasets (i.e., MNLI and QQP) and $16$ for single sentence datasets (i.e., SST-2). Also, the best learning rate has been determined through grid search among \{2e-5, 3e-5, 4e-5, 5e-5\}. 
We empirically selected the optimal value for $\alpha$ by a grid search between  0 and 1 with 0.1 increments. The best value of $\alpha$ is 0.6, 0.7, and 0.8 for SST-2, MNLI, and QQP, respectively. Finally, the masking percentage is set to 5\%, achieving the best performance on in-distribution and OOD  datasets. 
All experiments were performed with three epochs and an NVIDIA V100 GPU with five random seeds. The average accuracy value with corresponding standard deviations is reported on the in-distribution  and OOD test sets. We used huggingface code \cite{b42}
for performing the original Fine-tuning.

\section{Results and discussion}
\label{section:5}
Table \ref{tab:table2} displays the performance of Mask-tuning and other baselines, with some studies only reporting results for one model. We also compare the performance of Mask-tuning and original fine-tuning for the BART model, which has not been previously reported on OOD datasets. Finally, We present the results for GPT-3 on in-distribution and OOD dataset captured from \cite{b44}.\\
\textbf{Performance on OOD Datasets.} The results show that Mask-tuning performance outperforms all baseline models.  To the best of our knowledge, no study has reported BERT-base model performance on IMDB-Cont and IMDB-CAD. Thus, we compared the Mask-tuning and original fine-tuning performance.
In comparison with the BERT fine-tuning, Mask-tuning improves the performance on IMDB-Cont by +3.67\%, IMDB-CAD by +1.32\%, HANS by +12.62\%, AdvNLI by +2.2\%, and PAWS by +13.94\%. Moreover, in compare with the RoBERTa fine-tuning, Mask-tuning improves PLMs' performance on IMDB-Cont by +4\%, IMDB-CAD by +3.22\%, HANS by +7.9\%, AdvNLI by +6.2\%, and PAWS by +5.92\%. Finally, in compare with the BART fine-tuning, Mask-tuning enhances the BART performance on IMDB-Cont by +0.52\%, IMDB-CAD by +1.8\%, HANS by +14.18\%, AdvNLI by +4.8\%, and PAWS by +13.44\%.\\
\textbf{ Performance on In-Distribution Dataset.} Maintaining high performance on in-distribution datasets while improving performance on OOD datasets is essential in evaluating an approach. Mask-tuning achieves state-of-the-art performance on OOD datasets while boosting  performance on in-distribution datasets. These results highlight the effectiveness of Mask-tuning in balancing in-distribution and OOD performance.\\
\begin{table*}[t]
\centering
\caption{ Performance comparison of Mask-tuning (ours) and our ablation studies on various fine-tuning process.}
\resizebox{15cm}{!}{
\begin{tabular}{l|cccccccc}
\hline
\multicolumn{1}{l}{} &\multicolumn{3}{c}{Sentiment}& \multicolumn{3}{c}{NLI}  &\multicolumn{2}{c}{Paraphrase}\\\hline
\multicolumn{1}{l}{\textbf{}} &\multicolumn{1}{l}{In-distribution}&\multicolumn{2}{c}{OOD}& \multicolumn{1}{c}{In-distribution.} & \multicolumn{2}{c}{OOD}& \multicolumn{1}{c}{In-distribution}&\multicolumn{1}{c}{OOD}\\\hline
\multicolumn{1}{l|}{\textbf{Model}} &\multicolumn{1}{c|}{SST-2} &\multicolumn{1}{l|}{IMDB-Cont.} & \multicolumn{1}{l|}{IMDB-CAD} &\multicolumn{1}{c|}{\textbf{MNLI (m/mm)}} & \multicolumn{1}{c|}{\textbf{HANS}} &\multicolumn{1}{l|}{\textbf{AdvNLI}} &\multicolumn{1}{c|}{\textbf{QQP}}& \multicolumn{1}{c}{\textbf{PAWS}}\\\hline
\multicolumn{1}{l}{\textbf{BERT$_{base}$}} & \multicolumn{1}{c}{} & \multicolumn{1}{c}{} &\multicolumn{1}{c}{}&\multicolumn{1}{c}{}&\multicolumn{1}{l}{} & \multicolumn{1}{l}{} &\multicolumn{1}{l}{} \\\hline 
Fine-tuning on original train set   & 92.43& 79.08&\multicolumn{1}{c|}{87.00} &84.30/83.40& 56.90& \multicolumn{1}{c|}{24.12}& 90.80 & 32.80 \\
Fine-tuning on augmented train set.   &93.11 & 80.27&\multicolumn{1}{c|}{86.49} &84.06/84.28 &57.00 & \multicolumn{1}{c|}{25.70}&91.14  &33.93 \\
Mask-tuning w/o integrated loss  & 92.07&79.00 & \multicolumn{1}{c|}{84.99}& 83.57/84.22 &50.51&\multicolumn{1}{c|}{23.60} &89.92 &38.15 \\
Sequential Training  & 91.97&80.97 &\multicolumn{1}{c|}{86.76} & 83.40/84.00& 55.60&\multicolumn{1}{c|}{24.20} &91.00&34.82 \\ \hline
Mask-tuning (ours)  &\textbf{93.11}$\pm${0.1} &\textbf{82.75}$\pm${0.3} & \multicolumn{1}{c|}{\textbf{88.32}$\pm${0.1}}&\textbf{84.75}$\pm${0.2}/\textbf{85.10}$\pm${0.1} &\textbf{69.52}$\pm${0.2} &\multicolumn{1}{c|}{\textbf{26.32$\pm${0.3}}} &\textbf{91.54}$\pm${0.1}& \textbf{46.74}$\pm${0.5} \\\hline
\multicolumn{1}{l}{\textbf{RoBERTa$_{base}$}} & \multicolumn{1}{c}{} & \multicolumn{1}{c}{\textbf{}} & \multicolumn{1}{c}{\textbf{}}& \multicolumn{1}{c}{\textbf{}}\\\hline 
Fine-tuning on original train set  & 94.49&84.50 & \multicolumn{1}{c|}{88.40}&87.60/87.50& 67.80&\multicolumn{1}{c|}{31.20}&91.50& 38.45\\
Fine-tuning on augmented train set.   & 93.46& 85.54&\multicolumn{1}{c|}{89.76} &87.94/87.41 &72.66 & \multicolumn{1}{c|}{32.20}&91.26  &42.06 \\
Mask-tuning w/o integrated loss & 93.00& 81.80& \multicolumn{1}{c|}{87.31}&86.88/87.24 &65.16& \multicolumn{1}{c|}{30.00}& 90.09& 41.00\\
Sequential Training  &93.26 & 84.86& \multicolumn{1}{c|}{89.00}&87.50/87.40 &71.69& \multicolumn{1}{c|}{30.70}& 91.20&39.91 \\ \hline
Mask-tuning (ours)  &\textbf{94.60}$\pm${0.1} & \textbf{88.50}$\pm${0.2}& \multicolumn{1}{c|}{\textbf{91.62}$\pm${0.1}}& \textbf{87.72}$\pm${0.1}/\textbf{87.83}$\pm${0.2}&\textbf{75.70}$\pm${0.2}& \multicolumn{1}{c|}{\textbf{37.40}$\pm${0.6}}&\textbf{91.62}$\pm${0.2} &\textbf{44.37}$\pm${0.4}  \\ \hline
\multicolumn{1}{l}{\textbf{BART$_{base}$}} & \multicolumn{1}{c}{} & \multicolumn{1}{c}{} &\multicolumn{1}{c}{}&\multicolumn{1}{c}{}&\multicolumn{1}{l}{} & \multicolumn{1}{l}{} &\multicolumn{1}{l}{} \\\hline
\multicolumn{1}{l|}{Fine-tuning on original train set } & \multicolumn{1}{c}{93.23} & \multicolumn{1}{c}{82.48} &\multicolumn{1}{c|}{86.03}&\multicolumn{1}{c}{84.60/84.80}&\multicolumn{1}{c}{56.30} & \multicolumn{1}{c|}{30.51} &\multicolumn{1}{c}{90.50}&\multicolumn{1}{c}{32.27} \\
Fine-tuning on augmented train set.   &93.11 &82.99 &\multicolumn{1}{c|}{86.63} &85.90/86.00 &65.60 & \multicolumn{1}{c|}{32.80}&  9.60& 33.42\\
Mask-tuning w/o integrated loss  & 93.00& 82.02& \multicolumn{1}{c|}{85.80}& 84.02/84.10 &55.12&\multicolumn{1}{c|}{32.40} &89.34 & 35.21\\
Sequential Training  & 92.88& 82.65&\multicolumn{1}{c|}{86.13} &84.81/84.25 & 58.00&\multicolumn{1}{c|}{32.50} &89.95& 34.62\\ \hline
Mask-tuning (ours)  &\multicolumn{1}{c}{\textbf{93.80}$\pm${0.1}} &\multicolumn{1}{c}{\textbf{83.00}$\pm${0.1}} & \multicolumn{1}{c|}{\textbf{87.83}$\pm${0.3}}&\multicolumn{1}{c}{\textbf{86.08}$\pm${0.2}/\textbf{86.12}$\pm${0.3}} &\multicolumn{1}{c}{\textbf{70.48}$\pm${0.5}} &\multicolumn{1}{c|}{\textbf{35.31}$\pm${0.4}} &\multicolumn{1}{c}{\textbf{91.03}$\pm${0.1}}&\multicolumn{1}{c}{\textbf{45.71}$\pm${0.5}}  \\\hline
\end{tabular}}
\label{tab:table3}
\end{table*}\textbf{Comparing Mask-tuning and GPT-3 performance on OOD.}
\label{sub:C}
Table \ref{tab:table2} illustrates that GPT-3 (175B parameters) with few-shot prompting \cite{b44} achieves better accuracy on OOD data compared to BERT-based PLMs. However, this performance improvement was gained at the cost of a significant drop in accuracy on in-distribution downstream tasks' datasets. Our experiments show that Mask-tuning improves the generalization of PLMs by modifying the fine-tuning process. Mask-tuning achieves comparable or even better performance with GPT-3 on benchmark tasks like HANS in RoBERTa$_{base}$ (355M parameters), a larger pre-trained model than BERT$_{base}$ (110M parameters), without sacrificing performance on in-distribution data. These findings suggest promising avenues for future research in enhancing the fine-tuning process of PLMs to boost their generalization.\\
\textbf{ Analysing the impact of Mask-tuning on training examples' diversification.}
\label{sub:D}
Recent data augmentation studies \cite{b20, b22} showed that MLM improved PLM's generalization performance due to generating plausible and diversified examples from the original training examples. In this section, we compare the capabilities of MLM and Mask-tuning in creating plausible and diversified examples through three training epochs using BERT. We define a perturbed example as plausible if there is no syntactic or grammatical error after predicting a masked token. For example, if a NOUN is predicted for a masked VERB in an example that consequently makes the example grammatically incorrect, then we consider it as an implausible example. We randomly selected 200 training examples from each of the three in-distribution training datasets (SST-2, MNLI, and QQP). For each dataset, we ran two experiments: using 1) MLM  and 2) Mask-tuning for generating perturbed examples. Then we recorded the masked and predicted tokens for each example through three training epochs. We presented original and perturbed examples to 
two annotators to annotate them as plausible or implausible. Annotators ended up with 93\% inter-annotator agreement. For the remaining 7\%, the annotators expressed their views on each case in the presence of the researchers and finally 100\% agreement was achieved.  

We observed that in the first experiment (using MLM), MLM predicts a token precisely the same as the masked one in 76\% of examples. In 18\% of examples, MLM predicts a related plausible token to the masked one at least in one of the three epochs. In 5\% of examples, MLM predicts an implausible token and finally removes the masked token in 1\% of examples ([UNK]). 
However, the perturbation results when using Mask-tuning are entirely different from when using only the MLM.
In Mask-tuning, in 60\% of examples, the predicted token is a related plausible token to the masked one versus 18\% in the case of using only MLM. In 37\% of examples, mask-tuning predicts a token precisely the same as the masked one. In 2\% of examples, Mask-tuning predicts an implausible token and finally removes the masked token in 1\% of examples ([UNK]). 
This analysis shows that mask-tuning improves MLM's capabilities in generating plausible and diversified examples by more than three times. Hence,  it can be considered one of the main reasons for Mask-tuning to improve the PLMs' generalization performance. Table \ref{tab:table4} presents some perturbed examples generated by Mask-tuning.

\section{Ablation Study}
\label{sec:VI}
Three ablation experiments were performed to show the effectiveness of each component of Mask-tuning. 
We compare the Mask-tuning performance and ablation study on various fine-tuning process: data augmentation, Mask-tuning without integrated loss, and sequential training. Details of these experiments are as follows:\\ 
\textbf{Mask-tuning Vs. Data augmentation}. In this experiment, we aim to compare Mask-tuning performance and data augmentation. For this aim, we first used MLM to generate the perturbed examples from the original downstream task's training dataset. Then we fine-tuned the PLMs on the augmented training dataset, which is created by merging original and perturbed downstream training datasets. As we can see in Table \ref{tab:table3}, results show that directly augmenting the training data marginally improved the PLM's performance on both in-distribution and OOD datasets. However, this improvement is far behind the Mask-tuning performance. \\ 
\textbf{Mask-tuning without integrated-loss.} In this experimental setup, we investigate the effect of an integrated loss on the proposed method's performance. We train Mask-tuning on in-distribution downstream datasets based on only fine-tuning loss function output. Then evaluate the model on the in-distribution and the OOD evaluation set. As we can see from Table \ref{tab:table3} (third row of each model's results), eliminating MLM loss from the mask-tuning loss hurts the Mask-tuning performance on both in-distribution and OOD datasets. Hence, the integrated loss has an essential role in Mask-tuning performance and improving the PLMs' generalization. \\ 
\textbf{Mask-tuning Vs. Sequential training.} This experiment compares the performance of the Mask-tuning and sequential training on downstream tasks. For this comparison, we first use MLM to train the PLM  on the original downstream task training dataset. Then we tune the PLM by running original fine-tuning on the perturbed examples. Finally, we evaluated the PLM on in-distribution and OOD evaluation sets. The results in Table \ref{tab:table3} (fourth row in each PLM's results) show that sequential training noticeably decreased the PLMs' performance in most cases. 

All in all, the results of these experiments demonstrate the significance of integrating the three components of Mask-tuning (i.e., input perturbation, fine-tuning classification, and integrating loss) to improve the PLMs' generalization. 
 
 

\section{Conclusion}
In this study, we proposed Mask-tuning, a novel and effective training approach that addresses the challenges in PLMs' learning process, which affect their reusability on unseen data. By integrating the MLM training objective into fine-tuning, Mask-tuning enhances the PLMs' generalization. Our comprehensive experiments on in-distribution and OOD datasets from three downstream tasks on three PLMs showed that Mask-tuning outperforms six baselines on OOD datasets while also boosting PLMs' performance on in-distribution datasets. These findings demonstrate that Mask-tuning improves the practicality and effectiveness of PLMs for real-world applications. Moreover, since Mask-tuning is applicable to any PLM that works with fine-tuning, it has broad applicability. 


\vspace{12pt}

\end{document}